\documentclass[journal]{IEEEtran}

\usepackage{algorithm}
\usepackage{algorithmicx}
\usepackage[noend]{algpseudocode} 
\usepackage{acro}
\usepackage{amssymb}
\usepackage{amsmath}
\usepackage{booktabs} 
\usepackage{epsfig}
\usepackage{float}
\usepackage{graphicx}
\usepackage{mathtools}
\usepackage{multirow}
\usepackage{multicol}
\usepackage{xcolor}
\usepackage{times}

\DeclarePairedDelimiter\norm{\lVert}{\rVert}

\newcommand{\Fig}{Fig.}
\newcommand{\Sec}{Sec.}
\newcommand{\Tab}{Tab.}
\newcommand{\PathoSmall}{Patho\footnotesize{-200}\normalsize}
\newcommand{\PathoLarge}{Patho\footnotesize{-1000}\normalsize}

\newcommand{\etal}{\textit{et al}.}
\newcommand{\ie}{\textit{i}.\textit{e}.,}
\newcommand{\eg}{\textit{e}.\textit{g}.}

\DeclareAcronym{ROI}{
short=ROI,
long=region of interest,
}

\DeclareAcronym{ICC}{
short=ICC,
long=intrahepatic cholangiocarcinoma,
}
\DeclareAcronym{HCC}{
short=HCC,
long=hepatocellular carcinoma,
}

\DeclareAcronym{CT}{
short=CT,
long=computed tomography,
long-plural-form = computed tomographies,
}

\DeclareAcronym{MRI}{
short=MRI,
long=magnetic resonance imagery,
}

\DeclareAcronym{PET}{
short=PET,
long=positron emission tomography,
}

\DeclareAcronym{QA}{
short=QA,
long=quality assurance,
}

\DeclareAcronym{FCN}{
short=FCN,
long=fully convolutional network,
}

\DeclareAcronym{CNN}{
short=CNN,
long=convolutional neural network,
}

\DeclareAcronym{PTS}{
short=PTS,
long=primary tumor selection,
}

\DeclareAcronym{SaDT}{
short=SaDT,
long=spatially adaptive deep texture,
}

\DeclareAcronym{MIL}{
short=MIL,
long = Multi-instance Learning,
}

\DeclareAcronym{GPU}{
short=GPU,
long = Graphics Processing Unit,
}

\DeclareAcronym{KSF}{
short=KSF,
long = key slice filtering,
}

\begin{document}

\title{Harvesting, Detecting, and Characterizing Liver Lesions from Large-scale Multi-phase CT Data via Deep Dynamic Texture Learning}

\author{Yuankai~Huo,
        Jinzheng~Cai,
        Chi-Tung~Cheng,
        Ashwin~Raju,
        Ke~Yan,
        Bennett~A.~Landman,
        Jing~Xiao,
        Le~Lu,
        Chien-Hung~Liao,
        Adam~P.~Harrison
\thanks{J. Cai, K. Yan, L. Lu, A.P. Harrison are with PAII Inc., Bethesda, MD, USA; Y. Huo and A. Raju conducted this work while at PAII Inc., but are now with Vanderbilt University, Nashville, TN, USA and the University of Texas, Arlington, USA, respectively. (e-mail: huoyuankai@gmail.com; adampharrison070@paii-labs.com)}
\thanks{C. Cheng, C. Liao are with the Chang Gung Memorial hospital, Linkou, Taiwan, ROC.}
\thanks{J. Xiao is with PingAn Technology, Shenzhen, China}
\thanks{B.A. Landman is with the Department of Electrical Engineering and Computer Science, Vanderbilt University, Nashville, TN, USA}
}


\maketitle

\begin{abstract}
Non-invasive radiological-based lesion characterization and identification, \eg{}, to differentiate cancer subtypes, has long been a major aim  to enhance oncological diagnosis and treatment procedures. Here we study a specific population of human subjects, with the hope of reducing the need for invasive surgical biopsies of liver cancer patients, which can cause many harmful side-effects. To this end, we propose a fully-automated and multi-stage liver tumor characterization framework designed for dynamic contrast \ac{CT}. Our system comprises four sequential processes of tumor proposal detection, tumor harvesting, primary tumor site selection, and deep texture-based tumor characterization. Our main contributions are that, (1) we propose a 3D non-isotropic anchor-free detection method for liver lesions; (2) we present and validate \ac{SaDT} learning, which allows for more precise characterization of liver lesions; (3) using a semi-automatic process, we bootstrap off of $200$ gold standard annotations to curate another $1001$ patients. Experimental evaluations demonstrate that our new data curation strategy, combined with the \ac{SaDT} deep dynamic texture analysis, can effectively improve the mean F1 scores by $>8.6\%$ compared with baselines, in differentiating four major liver lesion types. Our F1 score of (hepatocellular carcinoma versus remaining subclasses) is $0.763$, which is higher than reported human observer performance using dynamic \ac{CT} and comparable to an advanced magnetic resonance imagery protocol.  Apart from demonstrating the benefits of our data curation approach and physician-inspired workflow, these results also indicate that analyzing texture features,  instead of standard object-based analysis, is a promising strategy for lesion differentiation.

\end{abstract}


%
\IEEEpeerreviewmaketitle

\acresetall

\section{Introduction}
\IEEEPARstart{L}{iver} cancer is one of the most fatal cancers in the world. Identifying lesions is crucial, as treatment strategies vary widely depending on the cancer subtype. Liver lesions can be diagnosed non-invasively, using clinical patient meta-information coupled with diagnostic imaging modalities, \eg, ultrasound, multi-parametric \ac{MRI}, or multi-phase dynamic contrast \ac{CT}. \ac{CT} remains the most common choice due to its cost-effectiveness and fidelity~\cite{oliva2004liver,burrowes2017contrast}. However, patients with unconfirmed radiological diagnoses require subsequent invasive procedures, such as biopsies or surgery, which can cause hemorrhages, infections, or even death~\cite{grant1999guidelines}. This is a non-negligible threat to these patients. Thus, improved non-invasive quantitative imaging based tumor characterization is a crucial goal to reduce the uncertainty of diagnosis~\cite{bilello2004automatic,frid2018gan,diamant2015improved,adcock2014classification,gletsos2003computer,mougiakakou2007differential,liang2018combining,yasaka2018deep,chen2019cascade}.  In this paper, we propose a multi-stage liver lesion characterization framework that incorporates effective 3D tumor detection, primary tumor site selection, and a principled deep texture learning for modeling tumor appearance in multi-phase \ac{CT}. Our solution can run fully automatically or semi-automatically given a human-drawn \ac{ROI}.

\begin{figure}
\begin{center}
\includegraphics[width=0.98\linewidth]{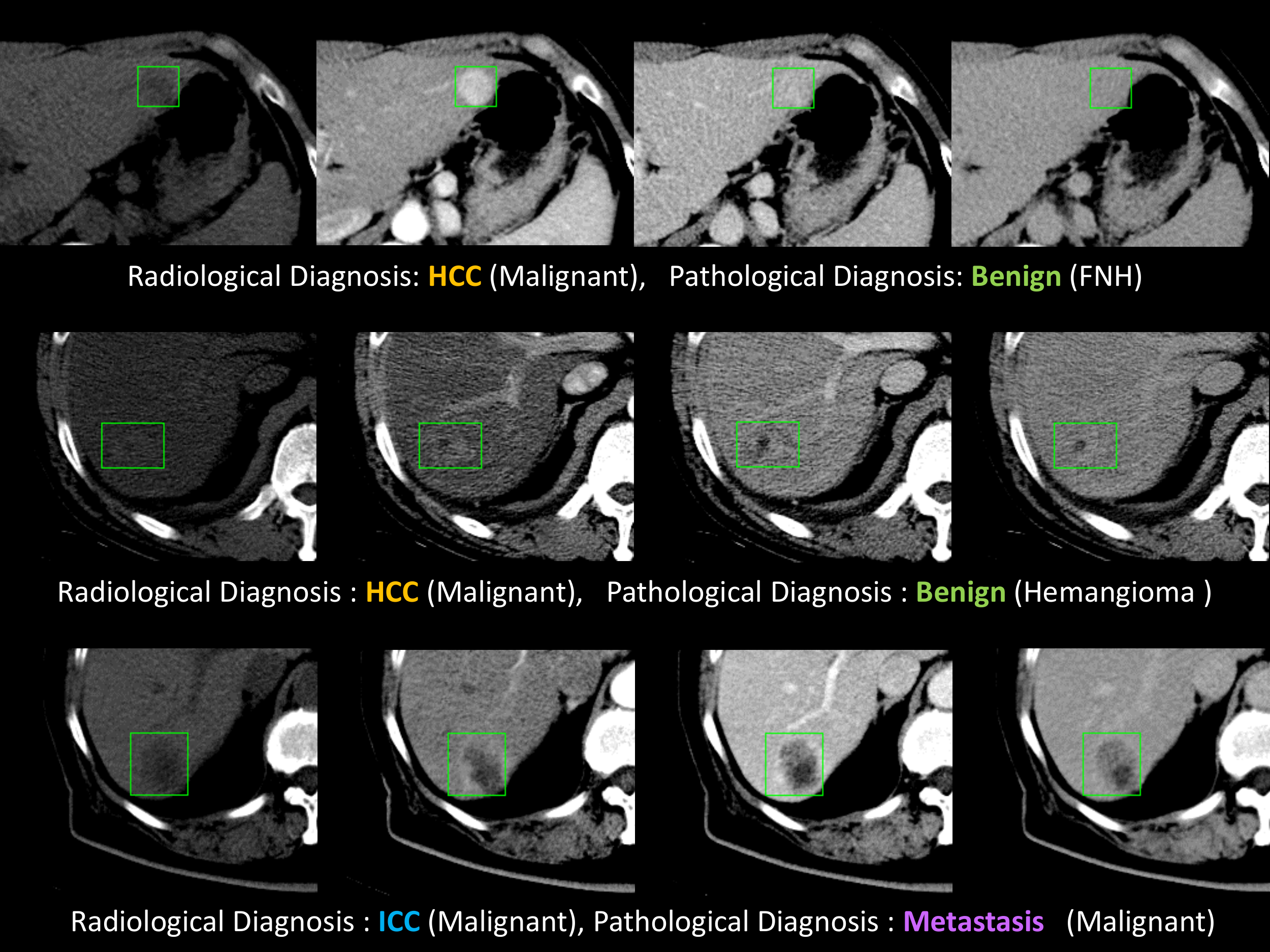}
\end{center}
   \caption{Difficult patient cases in our studied dataset where radiological diagnosis findings can be different from gold standard pathological ones. FNH is short for focal nodular hyperplasia. From left to right are non-contrast (NC), arterial (A), Venous (V), and Delayed (D) phase CT images.}
\label{fig:problem}
\vspace{-1em}
\end{figure}

\textbf{Why is this a difficult problem?} Although clinical diagnoses from multi-phase \acp{CT} are usually performed through consensus between radiologists and clinicians, misdiagnoses remain (\Fig~\ref{fig:problem}). For instance, benign and \ac{ICC} lesions can be misinterpreted as \ac{HCC}~\cite{freeman2006optimizing,kim2007peripheral}, with separating \ac{HCC} from hemangioma being a particularly pernicious difficulty~\cite{winterer2006detection,vilgrain2000imaging} . Solitary liver metastases are also difficult to differentiate from \ac{ICC}~\cite{kovavc2017intrahepatic} (\Fig~\ref{fig:problem}). Underscoring this, a 2006 retrospective study~\cite{freeman2006optimizing} investigating patients receiving a liver transplant to treat radiology-confirmed \ac{HCC}, discovered that $20\%$ of the treated tumors were in fact benign. Even with the same scoring systems, inter-reader variability between radiologist is still a problem for reproducible diagnosis~\cite{fowler2018interreader}. Khalili \etal{}~\cite{khalili2011optimization} report that F1 scores for physicians differentiating \ac{HCC} lesions from others is only $0.690$. We aim to leverage computer-aided approaches to reduce the uncertainties in diagnosis, with many of the above challenging cases. In this work, we aim to differentiate lesions into four subclasses (\ac{HCC} vs. \ac{ICC} vs. metastasis vs. benign), where even many human reader studies~\cite{Aube_2017,khalili2011optimization} only focus on binary classification (\ac{HCC} vs. others).

{\bf Method.}
Specifically, we (1) propose a fully automated end-to-end framework, and also (2) can train and test our approach on a scalable and partially-annotated dataset of 1305 multi-phase \ac{CT} studies/patients, all with histopathological ground truth. To accomplish this, we articulate several modules within the framework. A key module is a 3D generalization of the CenterNet model~\cite{zhou2019objects} to both curate our data and also detect liver tumors in deployment. Our data curation workflow boostraps off of existing large-scale public liver lesion cohorts~\cite{yan2018deeplesion,bilic2019liver}, along with $196$ manually annotated cases from our own dataset. This allows us to effectively exploit larger-scale hospital imaging data via semi-automated annotations. With all curated training data from both public and private sources in hand, we then execute classification in two stages. In the first stage, \ac{PTS} uses the aforementioned 3D tumor detector and a\ac{KSF}, trained using our curated data, to filter and select \ac{ROI} proposals for primary tumor slices. The second stage employs \ac{SaDT}, an analysis pipeline we develop to precisely model and differentiate the visual appearance on these primary tumor slices. This provides a determination of the lesion type. To the best of our knowledge, we are among the first studies to represent the task of radiological liver lesion characterization as a dynamic deep texture learning problem\footnote{As discussed in \cite{Aerts2018Artificial}, the primary imaging based cancer diagnosis cues are the potential tumor's size, shape and texture information.}.

\textbf{Experimental Challenges.} Our dataset consists of 1305 patients (5220 scans of four CT volumes per patient) with pathologically confirmed liver cancer labels. Importantly, our benign lesions represent some of the most challenging examples, \ie{} these are all lesions that ideally should not have been biopsied or resected in the first place, meaning that they are very hard and challenging to differentiate via \ac{CT} by human physicians. Validating our approach, we demonstrate that our data curation pipeline and deep dynamic texture learning method  can boost the four-class classification accuracy by $7.4$ and $3.7$ percent, respectively, compared to strong deep baselines. With the complete pipeline in place, we achieve a mean F1 score of $0.681$ in classifying \textbf{four} tumor subtypes and an F1 score of of $0.763$ for HCC vs. others. As we discuss later, this compares very favourably to physician performance, even when human readers only focus on the comparatively easier binary task of \ac{HCC} vs. others. As such, this work represents a significant step forward toward effective computer-aided diagnosis tools for liver lesion characterization via multi-phase CT imaging protocols.

\subsection{Related Work}

{\bf Automatic liver lesion characterization.} Liver biopsy remains the current gold standard to characterize the primary hepatic tumors (typically the largest). However, these are painful and risky procedures that can lead to serious complications. Therefore, in vivo liver tumor characterization technologies, \eg{}, medical imaging, have became an essential research direction to reduce the need for invasive assessments. Several works have proposed automatic imaging based lesion classifiers, with both hand-crafted~\cite{bilello2004automatic,diamant2015improved,gletsos2003computer,mougiakakou2007differential,adcock2014classification}, and deep features being used~\cite{frid2018gan,liang2018combining,yasaka2018deep,chen2019cascade}. Texture features have been shown to be beneficial~\cite{bilello2004automatic,gletsos2003computer,mougiakakou2007differential}, but prior to our work \emph{deep} texture features have not been investigated. For assessing performance, one must consider what liver lesion subtytpes are being distinguished. For instance, some only focus on distinguishing metastasis, cysts and hemangiomas (benign)~\cite{bilello2004automatic,frid2018gan,diamant2015improved}; or \ac{HCC}, cysts and hemangiomas~\cite{gletsos2003computer,mougiakakou2007differential}; or cysts, focal nodular hyperplasia (FNH) (benign), hemangioma, and \ac{HCC}~\cite{liang2018combining,chen2019cascade}. Cysts are typically very easy to identify by intensity levels alone, so we do not include them in our analysis, but instead focus on the more difficult problem of separating \ac{ICC}, \ac{HCC}, metastatic, and benign (hemangioma, FNH, and adenoma) lesions from each other. 

Patient selection is another factor that varies across studies. For example, some studies select only \ac{CT} studies confirmed by radiology alone~\cite{bilello2004automatic,adcock2014classification} and some by a mixture of clinical, radiological, and pathological diagnoses~\cite{diamant2015improved,gletsos2003computer, mougiakakou2007differential, yasaka2018deep}. Radiologically confirmed studies may not be well selecting for cases requiring the intervention of machine learning solutions~\cite{liang2018combining,chen2019cascade}, \ie{} the cases that are indeed difficult to diagnose via imaging. In contrast, using only pathologically confirmed studies can better represent these challenging cases, especially for the benign lesions (false positive cases) that underwent a biopsy or resection  procedure that ideally should have been avoided. Finally, apart from Chen \etal~\cite{chen2019cascade}, all previous approaches rely on manually drawn tumor \acp{ROI}. In this paper, we have developed an effective and automatic 3D primary tumor detection, selection and identification pipeline, and proposed a  multi-phase deep texture modeling system for precision tumor characterization, partially inspired by the clinical diagnosis workflow of human physicians.

{\bf Liver Lesion Detection.}
Within the deep learning era \cite{razzak2018deep}, liver tumor detection approaches can be categorized into bottom-up approaches~\cite{razzak2018deep,li2015automatic} or top-down ones~\cite{yan2018deeplesion,cai2018accurate}, with some recent work merging these two directions into a multi-task system~\cite{yan2019mulan}. Most prior work uses two-stage anchor based detection methods, \eg{}, derived from Faster-RCNN~\cite{ren2015faster} or Mask-RCNN~\cite{he2017mask}. Yet, recent advanced one-stage methods~\cite{zhou2019objects,law2018cornernet,lin2017focal,zlocha2019improving,cai2019one} have shown excellent performance while retaining noticeably simpler formulations (without the need to tune detection anchors for different applications). Among these, CenterNet~\cite{zhou2019objects} provides a solution well balanced in terms of complexity and performance. On the other hand, 2D detection methods can suffer from performance inconsistencies across CT slices and high false positive rates. Therefore, we employ CenterNet as our detection backbone, but extend it to 3D and tailor it to handle the spatially non-isotropic characteristics of radiology images.

{\bf Deep Dynamic Texture Learning.}
Texture analysis has been a canonical computer vision task for several decades. Traditional designs consist of three major steps: feature extraction~\cite{lowe2004distinctive,cula2001compact,leung2001representing}, dictionary based feature encoding~\cite{liu2011defense,fei2005bayesian,csurka2004visual}, and classification. Recently,  classic methods, \eg{}, Bag-of-words~\cite{csurka2004visual}, VLAD \cite{jegou2010aggregating}, and Fisher vectors (FV)~\cite{perronnin2010improving}, have been supplanted or augmented by deep texture learning ~\cite{cimpoi2015deep,zhang2017deep,andrearczyk2016using,liu2019bow}. Unlike the traditional texture analysis that has been widely used in medical imaging~\cite{castellano2004texture}, deep texture learning is not well explored, especially for tumor analysis. Given that texture is one of the most important visual features for radiologists~\cite{Aerts2018Artificial} for characterizing soft-tissue CT intensity patterns, texture analysis arguably should serve as a prominent focus for liver lesion characterization. To do this, we adapt and modify the recent DeepTEN network~\cite{zhang2017deep}, which provides an end-to-end \ac{CNN} method for texture classification. However, we introduce \acf{SaDT} to manage the physical meaning of \ac{CT} pixels and the varying sizes of lesions, which are not considerations within typical material/texture classification problems in computer vision. To the best of our knowledge, we are the first to exploit and employ deep dynamic texture learning for liver lesion classification.

\section{Methods}

\begin{figure*}[t]
\begin{center}
\includegraphics[width=0.85\linewidth]{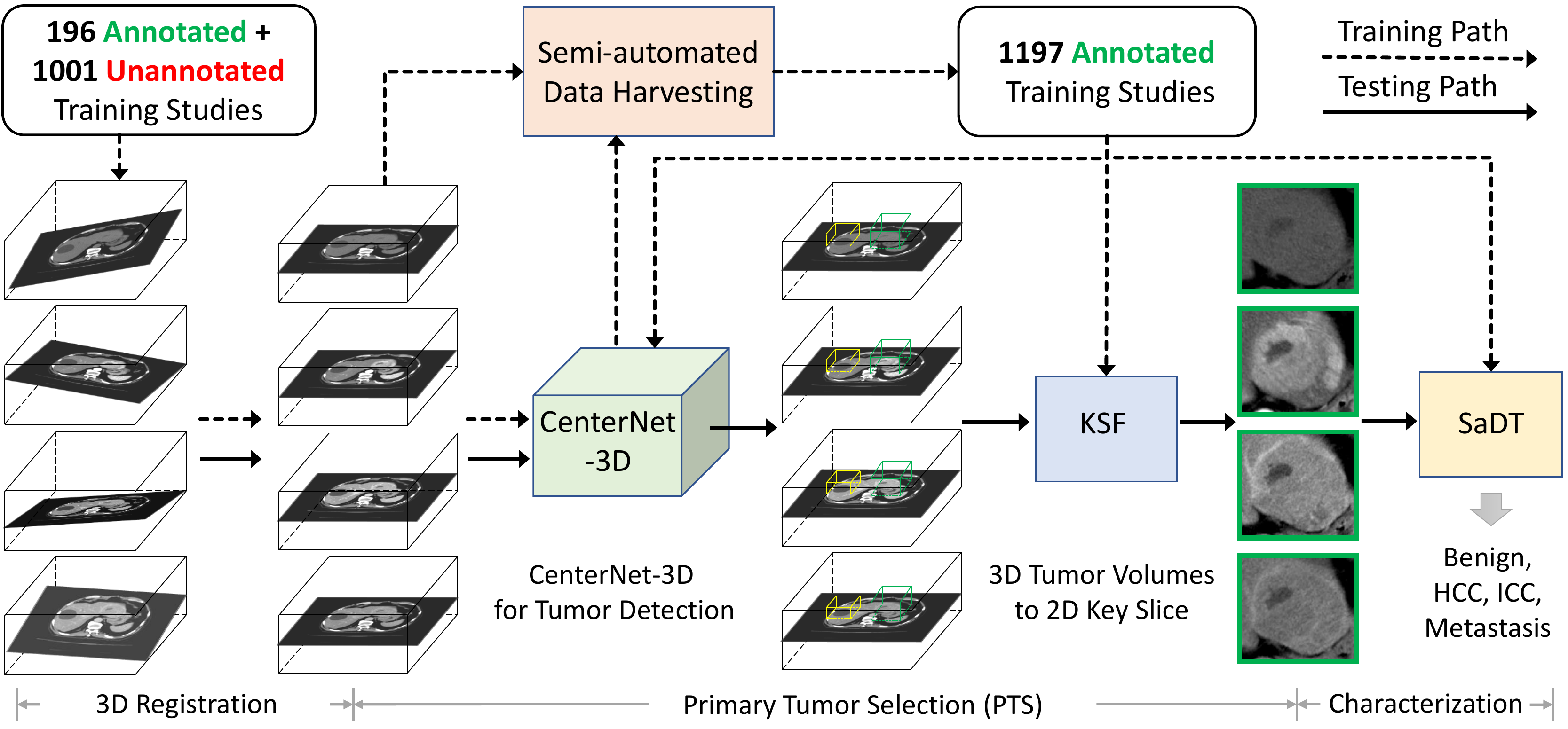}
\end{center}
   \caption{The proposed two-stage framework relies on \acf{PTS} and \acf{SaDT} classification, where the former, in turn, consists of detection and \acf{KSF}. The training and testing workflows are delineated using dashed and solid lines. 1001 previously unlabeled studies are curated by the semi-automated data harvesting. Then all 1197 annotated studies are used to train the proposed CenterNet-3D, \acs{PTS}, and \acs{SaDT} models.}
\label{fig:workflow}
\vspace{-1em}
\end{figure*}

\Fig~\ref{fig:workflow} illustrates our overall workflow. 
For pre-processing, we register all non-venous phase \ac{CT} scans of the same patient to the venous phase using DEEDS~\cite{heinrich2013mrf} and apply phase detection~\cite{Zhou2019Phase} to identify each scan's contrast phase. Our 3D tumor detection model  (\Sec~\ref{sec:3d_detection}) performs as the backbone module to facilitate both data harvesting (\Sec~\ref{sec:harvesting}) and \acf{PTS} (\Sec~\ref{sec:pts}), followed by using \ac{SaDT} to classify primary \acp{ROI} extracted by \ac{PTS} (\Sec~\ref{sec:texture}).

\subsection{Prerequisite: 3D Tumor Detection}
\label{sec:3d_detection}

\begin{figure}
\begin{center}
\includegraphics[width=0.8\linewidth]{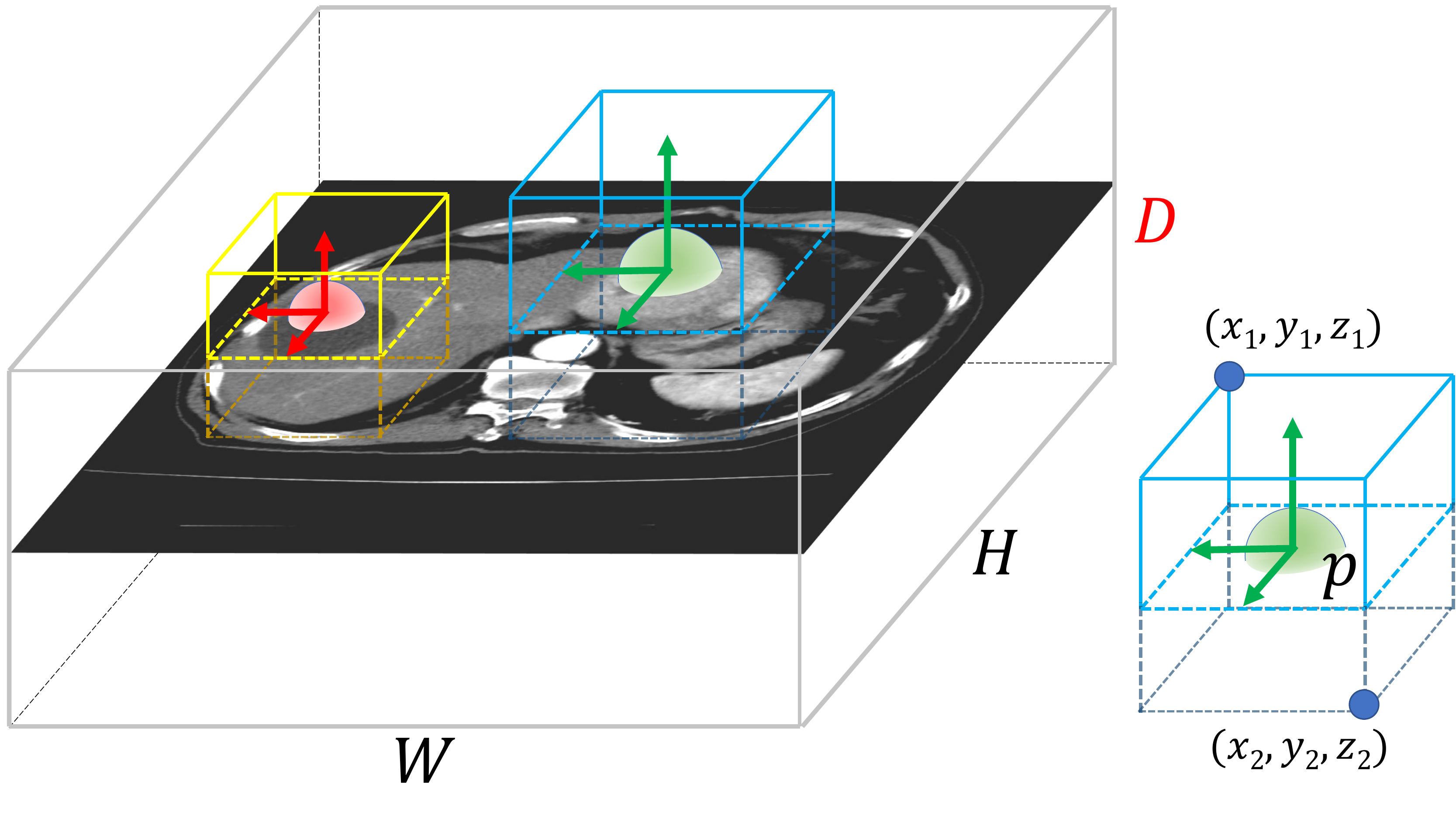}
\end{center}
   \caption{Liver lesion detection is performed using a 3D CenterNet-like detector. The red and green blobs are the 3D Gaussian kernels indicating the center points of two tumors. The lower-right subplot shows the 3D bounding box and a center point $p$. Note that the spatial resolution in the $D$ direction is typically lower than $W$ and $H$ resolutions in CT scans.}
\label{fig:detection}
 \end{figure}


As depicted in \Fig~\ref{fig:detection}, we opt for a 3D extension of 2D CenterNet detector~\cite{zhou2019objects}. The one-stage and anchor-free CenterNet based object detection models offers both high performance and simplicity, and does not require tuning anchor-based hyper-parameters required by two-stage and anchor-based approaches. We choose 3D detection  because the the intrinsic 3D structure of tumors may be too visually ambiguous when only observed from 2D slices~\cite{yan2019mulan}. 
CenterNet~\cite{zhou2019objects} follows a \ac{FCN} pipeline and the primary outcome is a 3D heatmap, $\hat Y \in [0,1]^{\frac{W}{R} \times \frac{H}{R} \times \frac{D}{R}}$, where $R$ is the downsampling factor of the prediction map. The heatmap should equal to $1$ at lesion centers and $0$ otherwise. We extend the stacked Hourglass-104 network~\cite{newell2016stacked} in the original CenterNet formulation~\cite{zhou2019objects} to a 3D variant. Following CornerNet~\cite{law2018cornernet} and CenterNet\cite{ zhou2019objects}, the ground truth of 3D target center points can be modeled as a 3D Gaussian kernel. Nevertheless since volumetric medical images are often non-isotropic, (\ie{}, the physical voxel spacing in $D$, \eg{}, 5 $mm$, is larger than that in $W$ and $H$, \eg{}, 1 $mm$), we use instead a non-isotropic Gaussian kernel:
\begin{equation}
{Y_{xyc} = \exp\left(-\frac{(\frac{x-\tilde p_x}{\gamma_x})^2+(\frac{y-\tilde p_y}{\gamma_y})^2+(\frac{z-\tilde p_z}{\gamma_z})^2}{2\sigma_p^2}\right)} \mathrm{,}
\end{equation} 
where $\tilde p_x$, $\tilde p_y$, and $\tilde p_z$ are the downsampled target center points ${\tilde p = \lfloor \frac{p}{R} \rfloor}$ and $\sigma_p$ is the kernel standard deviation. $\gamma_x$, $\gamma_y$, and $\gamma_z$ are the resolution coefficients to compensate for resolution differences. The corresponding pixel regression loss $L_{k}$, and the $\ell_1$-norm offset prediction loss $L_{off}$, are formulated identically as Zhou \etal~\cite{zhou2019objects}.

Given any 3D bounding box $(x_1, y_1, z_1, x_2, y_2, z_2)$, the center point is modeled as  $p = (\frac{x_1 + x_2}{2}, \frac{y_1 + y_2}{2}, \frac{z_1 + z_2}{2})$. The true bounding box size is computed as $\mathbf{s} = (x_2 - x_1, y_2 - y_1, z_2 - z_1)$. For a predicted bounding box $\hat{\mathbf{s}}$, the size $L1$ loss at the center point is calculated:
\begin{align}
    L_{size} = \frac{1}{N}\sum_{k=1}^{N} \norm{\hat{\mathbf{s}}_k - \mathbf{s}_k}_{1} \mathrm{.}
\label{eq:size_loss}
\end{align}
For generality, we employed the same hyper-parameter settings as ~\cite{zhou2019objects} to combine the three loss functions ($L_{k}$, $L_{off}$, $L_{size}$) and to set $\sigma_p$. To fit CT volumes into GPU memory, we first apply a robust multi-phase liver segmentation model~\cite{raju2020coheterogeneous}, crop around the resulting mask, and then resample the CT subvolume into 176$\times$256$\times$48.

\begin{figure}
\begin{center} 
\includegraphics[width=0.98\linewidth]{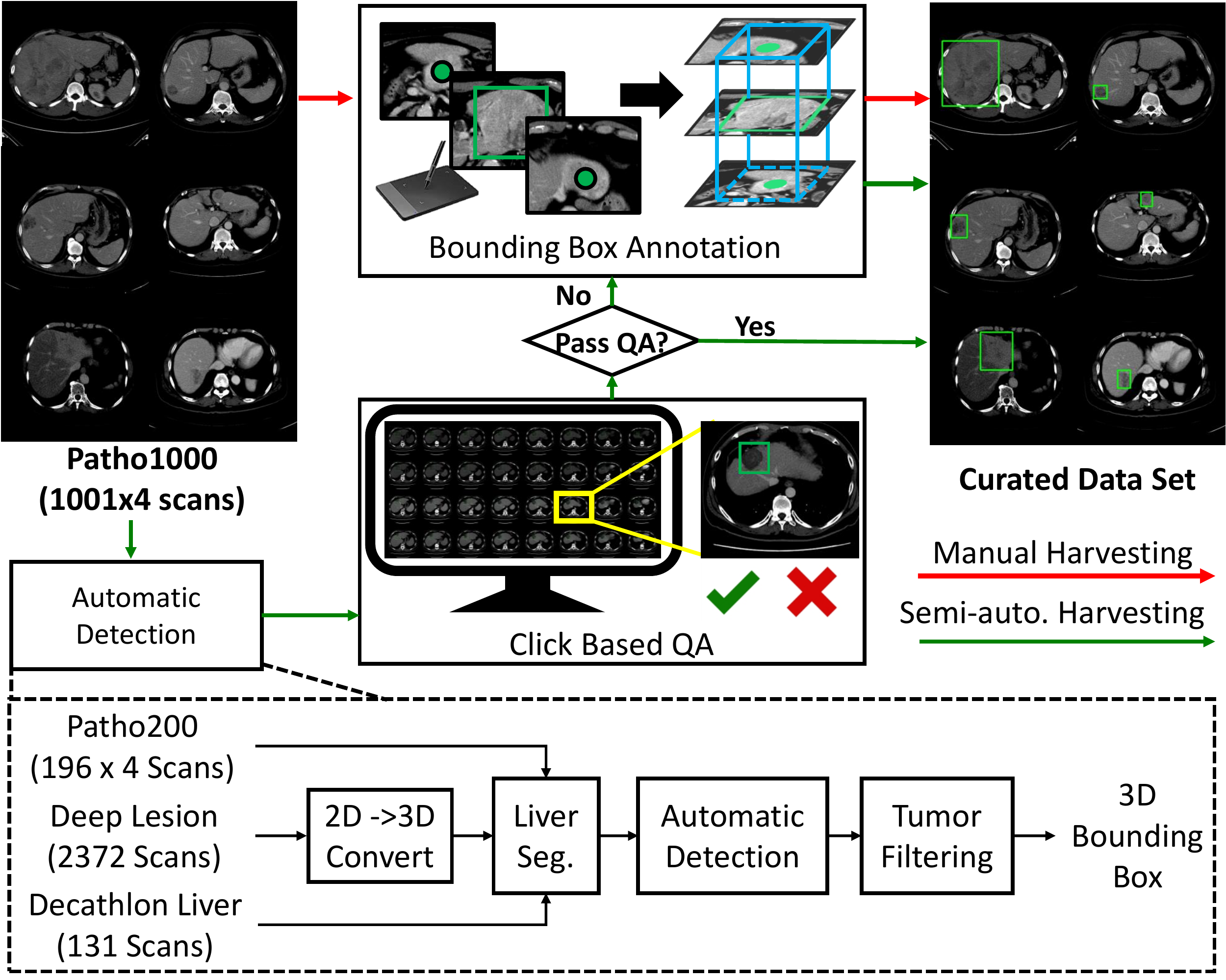}
\end{center}
\caption{The data harvesting framework uses a small cohort of bounding-box labelled data along with data collected from public sources~\cite{bilic2019liver,yan2018deeplesion}. For the latter, we select for liver lesions using the LesaNet semantic tags~\cite{yan2019holistic}. After training the detector we apply it to each multi-phase unlabelled volume. We render each candidate bounding box into a 2D image via a grid layout, which allows a human reader to rapidly categorized them into true/false positives using a mouse-click-based \ac{QA} process.}
\label{fig:data_curation}
\end{figure}

\subsection{Weakly Supervised Learning via Data Harvesting}
\label{sec:harvesting}
Ideally we would possess as large a dataset as possible of \ac{CT} volumes, all with bounding-box-labeled primary lesions. More typically though, one would only possess a small cohort of fully-labeled patient data,  $\mathcal{D}_{\ell}=\{g_{i},\ell_{i},\zeta_{i}\}_{i=1}^{N_{\ell}}$, where $\ell_{i}$ and $\zeta_{i}$ denote the image-level label of lesion type and bounding box or mask labels, respectively. It also typically possible to obtain an even larger cohort of patient data with only image-level labels, \eg{} data mined from hospital archives. We denote the weakly labelled data $\mathcal{D}_{u}=\{g_{i}, \ell_{i}\}_{i=1}^{N_{u}}$, where $N_{u}\gg N_{\ell}$. Based on our calculations, annotating 3D liver lesion bounding boxes consumes roughly $15$ minutes per \ac{CT} study (\Fig~\ref{fig:harvesting}) by a board-certified physician, making it prohibitive to completely annotate all patient studies.  Therefore we wish to develop scalable weakly supervised learning and data curation methods to deal with any large-scale dataset. 


\Fig~\ref{fig:data_curation} illustrates our approach. First, we collect or harvest a large collection of CT liver lesions from the public LiTS~\cite{bilic2019liver} and DeepLesion~\cite{yan2018deeplesion} datasets. For the latter, we select liver lesions by employing the LesaNet~\cite{yan2019holistic} produced semantic tags on identifying the lesion types. Using our own labelled multi-phase CT data, $\mathcal{D}_{\ell}$, and the single-phase public CT data~\cite{bilic2019liver,yan2018deeplesion}, we train our 3D CenterNet variant of \Sec~\ref{sec:3d_detection}. To harmonize the differing channel numbers of single (public) and multi-phase (private) CT data, we simply treat all CT phases as individual/independent observations. 

With a tumor detector trained, we apply it to $\mathcal{D}_{u}$. This provides us with a cohort of lesion candidates, but also a corresponding share of false positives. For each multi-phase CT volume $\in \mathcal{D}_{u}$, we obtain the tumor predictions from each phase, project them into the same volumetric space and merge the resulting candidates via non-maximal suppression \cite{cai2018accurate}. We then render each candidate bounding box, along with the corresponding multi-phase \ac{CT} intensities, into a 2D image representation using a grid layout. These 2D images can then be easily categorized into true or false positives using an efficient mouse-click-based \ac{QA} process. This takes on average one minute to verify all tumor candidates per study by a clinician, reducing the annotation labor by at least one order of magnitude. While this filters candidates into true and false positives, it is still possible to encounter studies where none of the detection candidates overlapped with the primary lesion(s) (for diagnosis). For these cases, they have to be mannually annotated from scratch. In our work, these constitute  $\sim20\%$ of  volumes in $\mathcal{D}_{u}$, meaning the labor required for annotation has still been considerably reduced. With this data curation pipeline in place, we can effectively and successfully harvest the liver tumor bounding boxes for the large-scale $\mathcal{D}_{u}$, which  significantly benefits both the downstream detection and classification performance. More importantly, our weakly supervised and human-in-the-loop data curation pipeline provides an annotated patient cohort with detected-verified/drawn tumor proposals we can use for the next process of primary tumor selection. 

\subsection{Primary Tumor Selection}
\label{sec:pts}

A fully automatic tumor characterization pipeline requires first finding the lesion(s) from the larger volume and then classifying them. Naturally, the 3D detector of \Sec~\ref{sec:3d_detection}, trained using the data curated by \Sec~\ref{sec:harvesting}'s processes, can localize lesions. However, additional filtering is necessary to extract reliable tumor \acp{ROI}. We present a dedicated \acf{PTS} module for this purpose, which is composed of the 3D CenterNet detector and a subsequent \acf{KSF} process.

\begin{figure}
\begin{center} 
\includegraphics[width=0.98\linewidth]{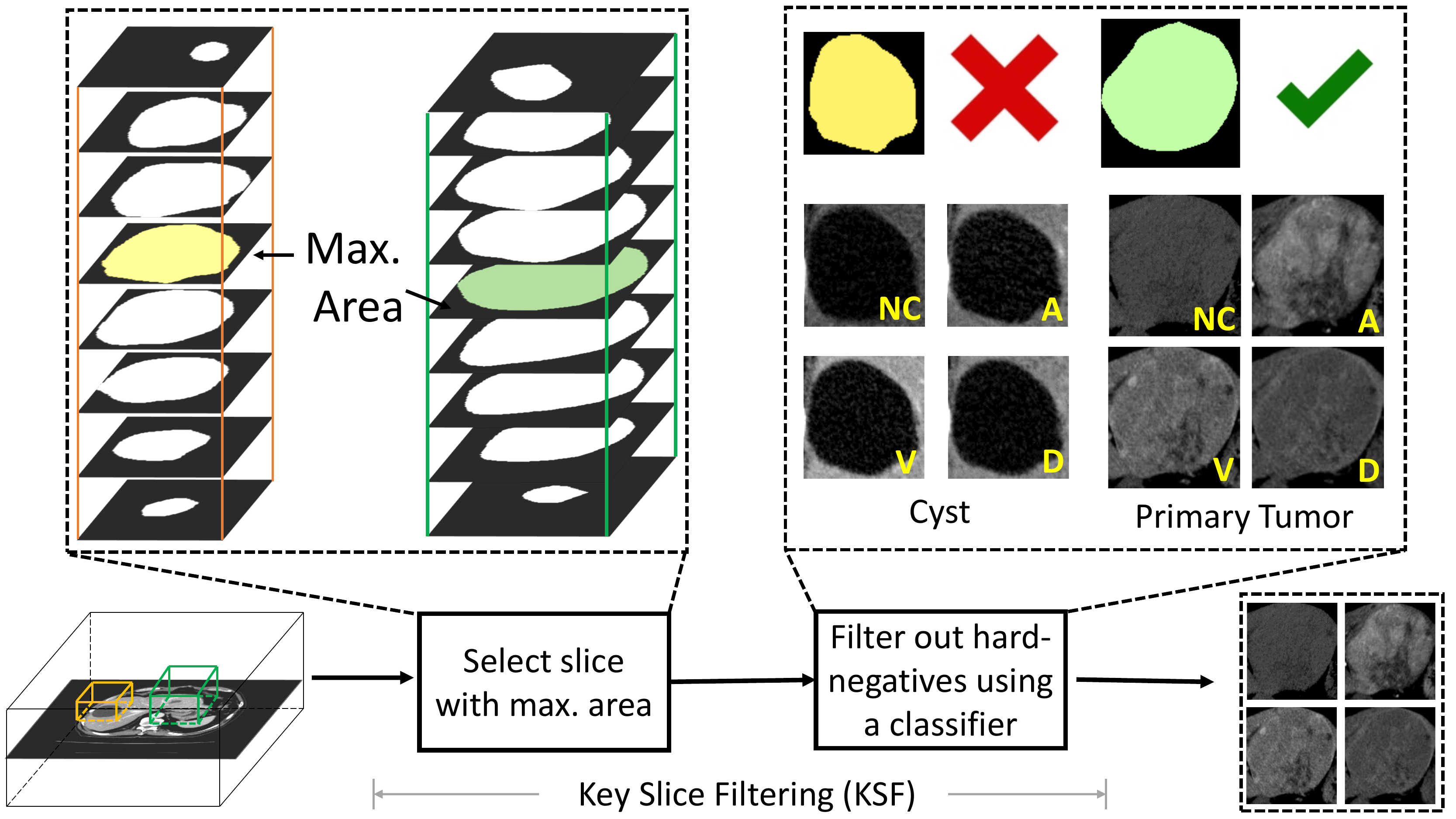}
\end{center}
   \caption{\acs{KSF} consists of choosing slices with the largest lesion mask area. Then non-primary key slices are filtered out by a binary classification network as hard-negative filtering. The filtered primary key slices are propagated to the next stage.}
\label{figure:pts}
\end{figure}

The first step of \ac{PTS} is to apply the 3D CenterNet detection to generate primary lesion candidates. We train this implementation using public datasets~\cite{bilic2019liver,yan2018deeplesion}, our labeled dataset $\mathcal{D}_{\ell}$ and the new harvested lesions in $\mathcal{D}_{u}$ from \Sec~\ref{sec:harvesting}. From this detector, we extract up to $10$ high-scoring 3D bounding boxes per volume. One or more of these candidates ideally overlap with the primary lesion(s); whereas many likely overlap with false positives, \eg{}, blood vessels, or non-significant lesions, like cysts (\Fig~\ref{figure:pts}). These undesirable candidates must be rejected. 

For reasons that \Sec~\ref{sec:texture} will explain, and our subsequent results would support, the downstream tumor characterization procedure is performed on 2D tumor \acp{ROI} across four \ac{CT} phases. Thus, any lesion candidate filtering must both select key 2D slices and, from these, select the most likely one belonging to a primary tumor. We call this \acf{KSF}. As \Fig~\ref{figure:pts} illustrates, we perform \ac{KSF} by first applying a binary lesion segmentation network, trained using $\mathcal{D}_{\ell}$, on each whole slice of the volume. Then for each 3D candidate, we choose the slice corresponding to the greatest prediction area as the key slice. While in principle any segmentation network can be incorporated, we choose to fine-tune the 2.5D segmentation head of the MULAN lesion detector~\cite{yan2019mulan}. We opt for this model for two reasons: (1) The segmentation head is pre-trained on DeepLesion~\cite{yan2018deeplesion} and should possess features with high lesion affinity; (2) the 2.5D segmentation allows us to incorporate sufficient background context without downsampling and compromising the CT resolution. 

With key slices selected for the top $10$ 3D candidate proposals, \ac{KSF} then separates them into primary and non-primary tumors. To do this, we apply a classifier on the multi-phase key slices (\Fig~\ref{figure:pts}). Thanks to the data curation pipeline in \Sec~\ref{sec:harvesting}, we have ample true- and false-positive candidates to train such a classifier. In this way, the data curation and \ac{KSF} steps can be viewed as hard-negative mining and hard-negative filtering, respectively. Any high-performing classifier can be used for this step, \eg{}, well-established appearance based \acp{CNN} or even the texture-based classifier we outline next in \Sec~\ref{sec:texture}. Next, for the tumors that are classified as primary, the corresponding detection scores will be used to break the ties, where the one with the largest detection score will be used as the primary tumor for the following \Sec~\ref{sec:texture}.

\subsection{Deep Dynamic Texture based Tumor Classification}
\label{sec:texture}

\begin{figure}
\begin{center} 
\includegraphics[width=0.98\linewidth]{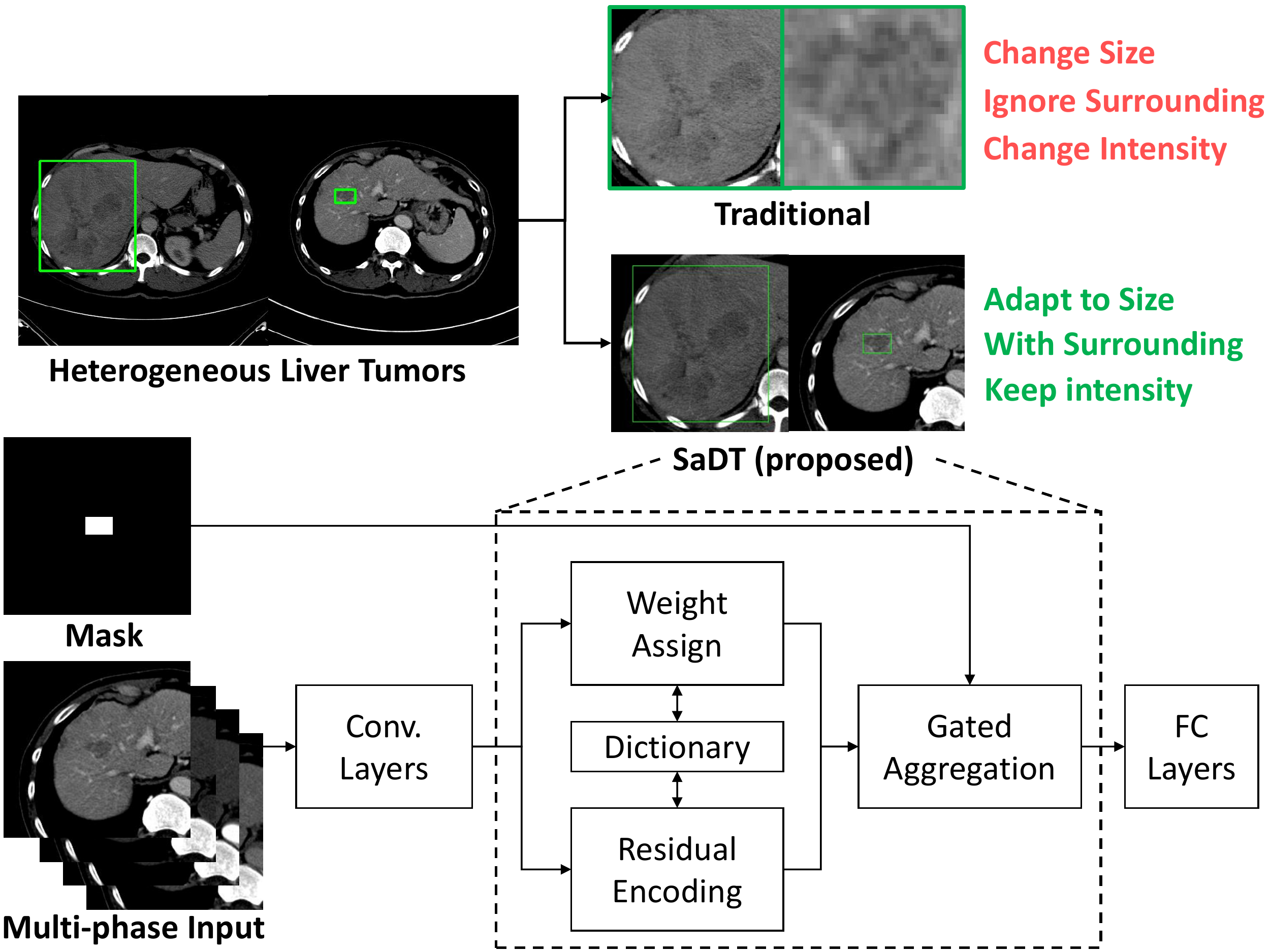}
\end{center}
   \caption{The proposed SaDT deep texture learning network. Convolution (Conv.) layers extract the image features, while the fully connected (FC) layers aggregate the final prediction. In the middle, our proposed SaDT method avoids encoding the spatial context outside the tumor masks (bounding boxes) to build more precise deep texture modeling and tumor appearance characteristics. }
\label{fig:texture}
\end{figure}

With the primary lesion \acp{ROI} localized and filtered using the \ac{KSF} of \Sec~\ref{sec:pts}, the next step is to differentiate them into \ac{HCC}, \ac{ICC}, metastasis, or benign (which includes hemangioma, focal nodular hyperplasia, and adenoma). Standard object recognition based \acp{CNN} can serve this function, but texture-based approaches have experienced success~\cite{gletsos2003computer,mougiakakou2007differential}, particularly prior to deep learning. In this work, we propose to formulate the tumor characterization problem (visually classifying liver tumor subtypes) using an unstructured deep texture approach. For physicians, texture features clinically serve as important cues for assessing soft-tissue properties~\cite{Aerts2018Artificial}, especially when observing the changing \ac{CT} intensities in the tumor site under different contrast phases. While it is theoretically possible to perform 3D texture modelling, the high inter-slice thickness of most \acp{CT}, \eg{}, $5$ mm in our data, is too coarse to capture texture. Instead, we apply a 2D deep texture learning workflow, adapting and enhancing the recent DeepTEN model~\cite{zhang2017deep} to create a \acf{SaDT} network. Using 2D models also allows us to use pretrained models, which is another key benefit. \Fig~\ref{fig:texture} depicts our \ac{SaDT} model.


Following DeepTEN~\cite{zhang2017deep}, texture modeling relies on counting, or soft-counting the codewords found in a set of visual descriptors. The visual descriptors, $F=\{f_1,...,f_M\}$, are generated from the features of an encoding \ac{FCN}, where $M$ is the number of spatial locations in the activation map. The residuals of each feature compared to a set of $K$ codewords $C=\{c_1, ..., c_K\}$ are then computed:
\begin{equation}
r_{ik}=f_i-c_k ,
\end{equation}
where $i=1,\ldots,M$ and $k=1,\ldots,K$. Classically these codewords have been fixed, but recent approaches allow these to be learned~\cite{zhang2017deep}. This is the approach we take.  

The set of all $M\times K$ residuals must now be aggregated into a global size $K$ feature  describing the overall texture. Before aggregation, the weights of each encoded residual are computed based on a ``soft-assignment''. Traditionally, only a single weight is assigned to each descriptor $x_i$, called ``hard-assignment'', but this makes the process non-differentiable. Therefore, the weight of each $r_{ik}$ is calculated using a softmax:
\begin{align}
\label{eq:assigning}
a_{ik}=
\frac{\exp(-s_k \|r_{ik}\|^2)}{\sum_{j=1}^K\exp(-s_j \|r_{ij}\|^2)} \mathrm{,}
\end{align}
where $s_k$ is the smoothing factor for each cluster center $c_k$. The smoothing factors are also designed as learnable parameters, which are updated during the training procedure.

Given the residuals as well as their weights, any arbitrary set of features, $F$, can be encoded into a fixed-size $K$ global vector:
\begin{equation}
e_k=\sum_{i=1}^M a_{ik}r_{ik} \mathrm{,}
\label{eq:aggregation}
\end{equation}
where all spatial locations are aggregated together. 

However, different from the standard texture learning scenarios, lesion sizes are heterogeneous across different cases (\Fig~\ref{fig:texture}). The simple resizing seen in most texture learners is not optimal, since it alters the intrinsic physical resolutions of the \ac{CT}. To address these spatial variations, we propose a new spatial adaptive aggregation that modifies \eqref{eq:aggregation} to
\begin{align}
e_k=\sum_{i=1}^M a_{ik}r_{ik}  \delta_i \mathrm{,}
\label{eq:adaptaggregation}
\end{align}
where each $\delta_i$ is a 0-1 binary value, indicating if the corresponding visual descriptor should be aggregated. In our implementation, $\delta_i$ is the tumor mask (bounding box) generated from detection \Sec~\ref{sec:pts}. Despite its straightforwardness, this process, denoted \ac{SaDT}, can result in significant performance improvements. Lastly, the final output vector is normalized using an $\ell_2$-norm. 
Since the training cohorts are highly unbalanced, for all experiments, the same weighted focal-loss ($\gamma$ = 2)~\cite{lin2017focal} is used as the classification loss function to reduce overfitting to the dominating class. The weights were set to $5$ for HCC, $2$ for metastasis, and $1$ for remaining classes. The same stochastic gradient descent optimizer with a learning rate of $0.004$ is used for all experiments. 

\setlength{\tabcolsep}{7pt}
\begin{table*}
\caption{The data used in this study include the publicly available LiTS and DeepLesion dataset. Patho is our in-house multi-phase dynamic CT dataset, which contains non-contrast (NC), arterial (A), venous (V), and delayed (D) phases.  Patho-200 and Patho-test are manually annotated, while Patho-1000 is annotated using our semi-automated data harvesting. All training data are used in detection and tumor harvesting, while only Patho-200 and Patho-1000 are used in tumor characterization.}
\centering
\begin{tabular}{l@{}c@{\ \ }ccccccc}
\toprule
& Dataset & Annotation & Studies & Phases & Categories \\
\midrule
\multirow{4}{0.65in}{Train \& Validation} & LiTS\cite{bilic2019liver} & 3D Seg. &  131 &  $V$  & Liver Tumor (131)\\
& DeepLesion (DL)\cite{yan2018deeplesion} & 2D RECIST & 2372 & $Unkown$ & Liver Lesion (2372) \\
& \PathoSmall & 3D Seg. & 196  & $NC, A, V, D$ & \ HCC(52), ICC(40), Benign(55), Meta(49)\\
& \PathoLarge & 3D B.Box & 1001  & $NC, A, V, D$ & HCC(921), ICC(4), Benign(2), Meta(74)\\
\midrule
\multirow{1}{0.65in}{Test} & Patho-test & 3D B.Box & 108  & $NC, A, V, D$ & HCC(60), ICC(8), Benign(16), Meta(24)\\

\bottomrule
\end{tabular}

\label{table:tab1}
\end{table*}

\setlength{\tabcolsep}{1.4pt}

\section{Experiments}

{\bf Data:} We collected multi-phase dynamic \ac{CT} scans from the \textit{Chung Gung Memorial Hospital, Taiwan, ROC}, selecting patients with liver lesions that received surgical resection or percutaneous biopsy in the period between June 2003 to April 2018. \ac{CT} scans were acquired from the hospital archives within one month before the invasive procedure. This resulted in $1305$ studies, with all \ac{CT} scans having standard $5$ mm slice thickness. Most patients underwent resection ($979$, $72.2\%$), with the remainder receiving biopsy ($377$, $27.8\%$) for pathological confirmation. The dataset was split into training ($1197$) and testing ($108$) studies, keeping the distribution of lesion types the same. From the training set, the lesions of $196$ studies were manually segmented under the supervision of a trained clinician (\PathoSmall), with the remainder only having image-level labels (\PathoLarge). This corresponds to $\mathcal{D}_{\ell}$ and $\mathcal{D}_{u}$, respectively. Because of the paucity of certain lesion types, during model development both \PathoSmall~ and \PathoLarge~ were split into 5-folds for cross validation for more stable assessments. \Tab~\ref{table:tab1} provides more data details. The average voxel size is $0.69\,\textrm{mm} \times 0.69\,\textrm{mm} \times 5\,\textrm{mm}$.

\begin{figure}
\begin{center} 
\includegraphics[width=0.8\linewidth]{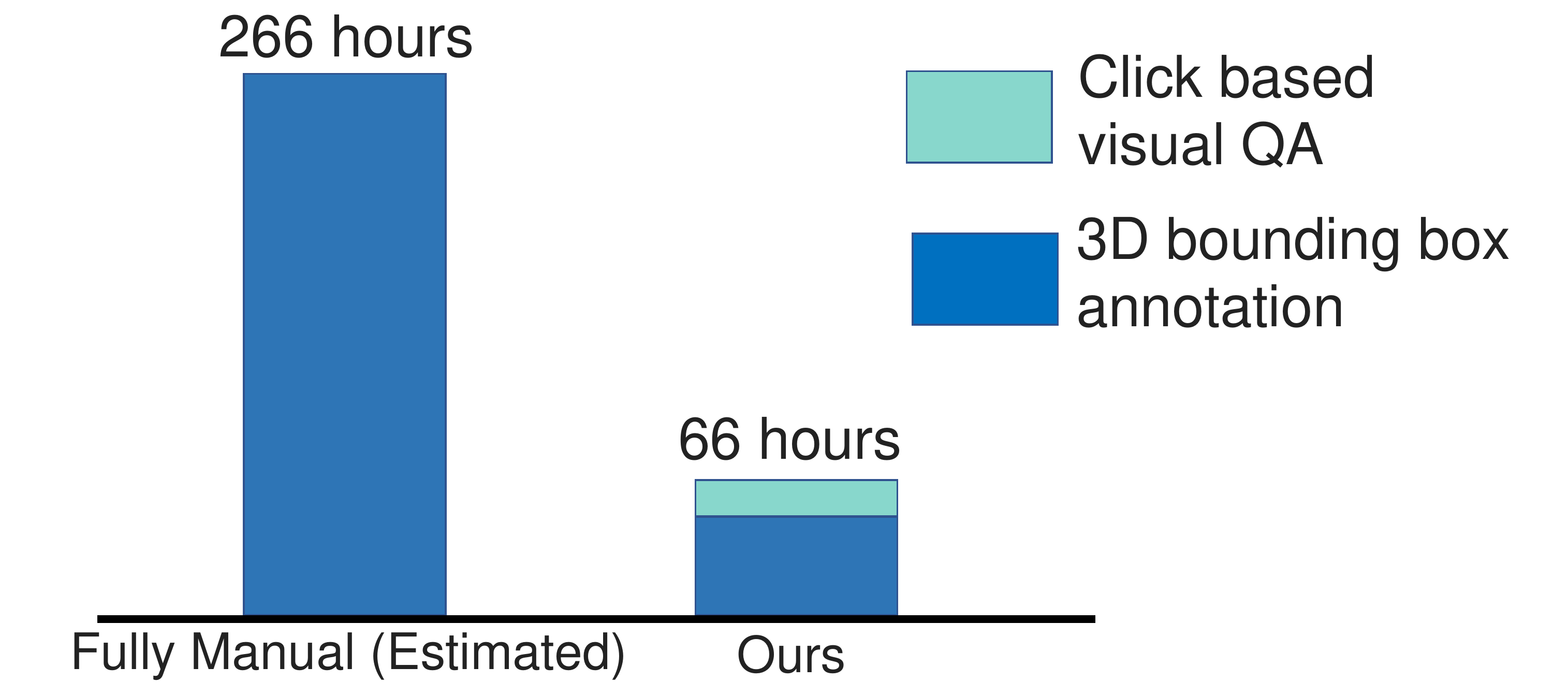}
\end{center}
\vspace{-1em}
\caption{Labor time using different curation strategies.}
\label{fig:harvesting}
\vspace{-1em}
\end{figure}

Since only \PathoSmall~ has localizations, we use the data curation pipeline of \Sec~\ref{sec:harvesting} to extract tumor \acp{ROI} from \PathoLarge. The labor savings ($75\%$) provided by the curation is presented in \Fig~\ref{fig:harvesting}. In total, this required employing our mouse-click \ac{QA} on $1064$ lesion candidates ($\approx1$ min/study) and executing 3D bounding box annotations ($\approx15$ min/study) on 193 patients whose primary tumors were not captured by the proposals. Our test set consists of $108$ patients ($432$ multi-phase CT scans), which were fully annotated by our collaborating physician with $15$ years of experience. This population is comparable or larger than previous studies~\cite{chen2019cascade,yasaka2018deep,liang2018combining}. 


\setlength{\tabcolsep}{3pt}
\begin{table*}
\caption{We measure lesion characterization using overall accuracy ($Acc.$), mean $F_1$ score, and  one-vs.\-all $F_1$ scores.}
\centering
\begin{tabular}{l@{}c@{\ \ }cccccccccccc}
\toprule
 &  & \PathoSmall & \PathoLarge & $Acc$. & mean $F_1$ & $F_1$(HCC) & $F_1$(ICC)& $F_1$(Benign)&	 $F_1$(Metastasis) 
  \\
\midrule
\multirow{2}{0.5in}{ 3D} & ResNet3D \cite{hara2017learning} & \checkmark &   &  0.333 & 0.267 & 0.451 & 0.091 & 0.270 & 0.255 \\
& ResNet3D \cite{hara2017learning} & \checkmark & \checkmark &  0.583 & 0.426 & 0.710 & 0.182 & 0.645 & 0.167 \\

\midrule

\multirow{8}{0.5in}{ 2D} & ResNet \cite{he2016deep} & \checkmark &  &   0.565& 0.486 & 0.691 & 0.250 & 0.625 & 0.379 \\
& ResNet \cite{he2016deep} & \checkmark & \checkmark   &  0.556 & 0.466 & 0.636 & 0.118 & 0.649 & 0.462 \\
& ResNet-Ensemble & \checkmark &  &  0.500 & 0.494  & 0.606 & 0.300 & 0.579 & 0.490 \\
& ResNet-Ensemble & \checkmark & \checkmark  &  0.694& 0.594 & 0.774 & 0.286 & \textbf{0.750} & 0.565 \\
& DeepTEN \cite{zhang2017deep} & \checkmark &   &  0.593 & 0.519 & 0.679 & 0.261 & 0.541 & 0.596 \\
& DeepTEN \cite{zhang2017deep} & \checkmark &  \checkmark & 0.639& 0.527 & 0.742 & 0.286 & 0.545 & 0.533 \\
& DeepTEN-Ensemble & \checkmark &   &  0.528 & 0.595  & 0.731 & 0.455 & 0.571 & 0.625 \\
& DeepTEN-Ensemble & \checkmark & \checkmark  & 0.685& 0.540 & 0.779 & 0.286 & 0.593 & 0.500 \\

\midrule
\multirow{4}{0.5in}{ 2D (Ours)} & SaDT & \checkmark &   & 0.620& 0.541& 0.710 & 0.286 & 0.526 & 0.640 \\
& SaDT & \checkmark & \checkmark  &   0.713 & 0.622 &	\textbf{0.800} &	0.421 &	0.615&	0.652& \\
& SaDT-Ensemble & \checkmark & &  0.657& 0.580   & 0.727 & 0.316 & 0.595 & 0.680 \\
& SaDT-Ensemble & \checkmark & \checkmark & \textbf{0.731}& \textbf{0.681} & 0.797 & \textbf{0.500} & 0.690 & \textbf{0.735}\\

\bottomrule

\end{tabular}
\vspace{0.5em}


\vspace{-0.5em}
\label{table:mainres}
\end{table*}
\setlength{\tabcolsep}{1.4pt}

\subsection{Tumor Characterization}
\label{sec:tumor_charac}
We first validate our choice of \ac{SaDT} model for lesion characterization, assuming for now that manually drawn \acp{ROI} are available. As comparison, we test against both 3D and 2D classification networks. For a fair comparison, we use an ImageNet pretrained ResNet-50 backbone for all 2D networks. For 3D networks, we test ResNet3D~\cite{hara2017learning,he2016deep}, while for 2D networks we test standard ResNet~\cite{he2016deep} and also the DeepTEN network~\cite{zhang2017deep}. The same single top-1 most-confident tumor (largest volume size) for a given scan is used to train both 3D and 2D classifiers. For 2D, the top-5 largest-area slices are employed to train the network, while the top-1 largest slices are used during testing. For 3D, the entire top-1 mask center cropped tumor volume is used for both training and testing. Such design ensures only one prediction per patient, which enables the validation using the image level ground truth.

For DeepTEN, the tumor region within the bounding box is resized to $256 \times 256$ following typical texture analysis practices~\cite{zhang2017deep}. For the \ac{SaDT} method, the bounding box is encoded as a binary mask channel for spatial adaptive aggregation (\Fig~\ref{fig:texture}). When compared against ResNet, DeepTEN reveals the impact of applying texture analysis vs.\ standard appearance-based \acp{CNN}. Comparisons against \ac{SaDT} reveal the impact of our spatially adaptive approach to deep texture learning. We also perform model ensemble experiments, which apply all five models trained in cross validation to the test set with majority voting. We do not perform ensembling using 3D methods due to the heavy computational burdens. However, when using single-model inference, we ensure that the same train-validation fold is used across all methods. We evaluate the methods using overall accuracy ($Acc.$) as well as one-vs.\-all F1 scores ($F_1$).

 \Tab~\ref{table:mainres} presents the results. Some immediate conclusions can be drawn. {\bf First}, the inclusion of \PathoLarge{} significantly boosts performance for most models, validating our semi-automatic data curation pipeline. {\bf Second}, 2D models tend to perform better than 3D variants. {\bf Third}, from the 2D models, texture-based variants outperformed the object appearance-based ones, demonstrating the value of applying multi-phase deep texture learning for lesion characterization. {\bf Fourth}, unsurprisingly, ensembling aided all models. {\bf Finally}, \ac{SaDT} achieves the highest, or nearly highest performance, across all metrics except the F1(benign). While 2D ResNet exhibits the highest performance for this metric, its other metrics are much worse, whereas \ac{SaDT} provides much more stable performance across lesion types. Thus, these results demonstrate the benefits of applying \ac{SaDT} to lesion analysis. Given that these include some of the most difficult cases to differentiate\footnote{This patient population represents some of the most challenging lesion characterization cases which cannot confidently be resolved using radiological imaging by human readers, even assisted with patient meta-information (e.g., blood work, age, etc.).}, these metrics demonstrate the promise of our lesion characterization strategy.  
 
 
\setlength{\tabcolsep}{2pt}
\begin{table}
\caption{Detection Results. The percentage of the patients with at least one primary tumor is detected (P1TD) is presented. P1TD-1 means only one most confident primary tumor detection is provided by the detector, while P1TD-10 indicates the scenarios that at most 10 confident tumor detection results are offered.}
\small
\centering
\begin{tabular}{l@{}c@{\ \ }ccccccc}
\toprule
& \PathoSmall & LiTS+DL & \PathoLarge & P1TD-1 & P1TD-10\\
\midrule

\multirow{2}{0.6in}{CenterNet -2D} & \checkmark & & & 0.741 & 0.929 \\
& \checkmark & \checkmark  && 0.714 & 0.929 \\
& \checkmark & \checkmark  & \checkmark  & 0.786 & \textbf{0.973} \\
\midrule
\multirow{3}{0.6in}{CenterNet -3D (Ours)} & \checkmark &  &  & 0.571 & 0.848 \\
& \checkmark & \checkmark && 0.732 & 0.929 \\
& \checkmark & \checkmark &\checkmark  & \textbf{0.795} & 0.938\\

\bottomrule
\end{tabular}
\vspace{0.5em}

\label{table:detection}
\end{table}
\setlength{\tabcolsep}{1.4pt}

\setlength{\tabcolsep}{5pt}
\begin{table}
\caption{The key slice filtering (KSF) performance.}
\centering
\begin{tabular}{l@{}c@{\ \ }ccccccc}
\toprule
& $Accuracy$ & $F_1$(Primary Tumor)  \\
\midrule
\multirow{1}{1in}{ResNet} & 0.877 &  0.888 \\
\multirow{1}{1in}{SaDT (Ours)} & \textbf{0.887}  & \textbf{0.891} \\

\bottomrule
\end{tabular}
\vspace{0.5em}

\label{table:ptf}
\end{table}
\setlength{\tabcolsep}{1.4pt}

\setlength{\tabcolsep}{5pt}
\begin{table*}
\caption{Tumor characterization of automatied pipelines. The best performance from \Tab~\ref{table:tab1} is shown as the upper bound performance.}
\centering
\begin{tabular}{l@{}c@{\ \ }cccccccccccc}
\hline
 &  $Acc.$ & mean $F_1$ &$F_1$(HCC) & $F_1$(ICC)& $F_1$(Benign)& $F_1$(Metastasis) 
  \\
\hline
\multirow{1}{2.5in}{Manual Detection + SaDT (Upper Bound)} & \textbf{0.731} & \textbf{0.681} &\textbf{0.797} & \textbf{0.500} & \textbf{0.690} & \textbf{0.735}\\
\hline
\multirow{1}{2in}{CenterNet-2D + SaDT} & 0.565 & 0.529 & 0.667 & 0.500 & \textbf{0.480} & 0.468 \\
\multirow{1}{2in}{CenterNet-3D + SaDT} & 0.639 &0.521 & 0.741 & 0.375 & 0.455 & 0.512 \\
\multirow{1}{2in}{CenterNet-2D + KSF + SaDT} &  0.639 & 0.538& 0.748 & 0.444 & 0.435 & 0.524 \\
\multirow{1}{2in}{CenterNet-3D + KSF + SaDT} &\textbf{ 0.657} & \textbf{0.583} & \textbf{0.763} & \textbf{0.533} & 0.476 & \textbf{0.558}\\
\hline
\end{tabular}
\vspace{0.5em}

\label{table:pipeline}
\vspace{-1em}
\end{table*}
\setlength{\tabcolsep}{1.4pt}

\subsection{Primary Tumor Selection}

While \Sec~\ref{sec:tumor_charac} demonstrates the value of our texture learning approach, \ac{SaDT} alone does not provide a complete pipeline. For this reason, we also validate our detection network. To do this, we test both the 2D CenterNet~\cite{law2018cornernet} as well as our proposed CenterNet-3D variant. For 2D detection, \ac{CT} slices are first normalized to $0.8\,\mathrm{mm}\times0.8\,\mathrm{mm}$ pixel size, and then either center cropped or zero padded to 512 $\times$ 512. Three consecutive slices (with $2$ mm slice thickness) are used as three input channels. We use the Hourglass~\cite{newell2016stacked} network, pretained on MS-COCO~\cite{lin2014microsoft}, as backbone. Thus the 2D CenterNet represents a strong baseline. For the CenterNet-3D, liver segmentation, cropping, and resizing is performed as described in \Sec~\ref{sec:3d_detection}. The hyper parameters of both CenterNet-2D~\cite{law2018cornernet} and the CenterNet-3D are optimized based on the validation dataset. Since the goal is to have the detector capture the primary tumor in its top $10$ candidates (followed by a classifier to further filter these candidates), we measure performance by the percentage of patients with at least one primary tumor detected (P1TD). This metric can be divided into whether the first $1$ (P1TD-1) or first $10$ (P1TD-10) candidates captured a primary tumor. P1TD-10 measures whether we are able to even capture a primary tumor in our candidate cases. But since we break ties in \ac{KSF} using detection scores, it is also important to measure P1TD. We test each variant when trained only on (A) \PathoSmall, (B) \PathoSmall~  + public data~\cite{bilic2019liver,yan2018deeplesion}, and (C) \PathoSmall~ + \PathoLarge~ + public data, with the latter measuring the impact of our data curation on detection performance.



\Tab~\ref{table:detection} presents the detection results. As can be seen, incorporating the harvested lesions benefits all models, further validating our data curation pipeline. Comparing the two variants, CenterNet-3D exhibits lower P1TD-10 accuracy while producing higher P1TD-1 accuracy. Given the complex interaction, it is difficult to assess which balance between the two metrics is best. However, as \Sec~\ref{sec:unified} will demonstrate, when incorporated within our complete automatic pipeline, CenterNet-3D corresponds to higher overall performance, making it our preferred choice. 

In addition to the detector, \ac{KSF} selects primary tumor slices from the detection candidates. A key component of \ac{KSF}, in turn, is a classification network to filter out spurious detections. As such, it is important to characterize this classifier performance. To do this, we measure the classifier performance on slices extracted from the test set, compared to a standard ResNet vs.\ our \ac{SaDT}. As can be seen in \Tab~\ref{table:ptf}, both models achieves near 90\% accuracy. However, because the \ac{SaDT} edges out standard ResNet, and we also employ the former for both \ac{KSF}, in addition to its main use within the downstream lesion classification.

\subsection{Fully Automatic Tumor Classification Framework}
\label{sec:unified}

Finally, we test the contribution of each component to a fully automatic pipeline that includes both the \ac{KSF} and \ac{SaDT} modules. We compare against an upper bound where lesion \acp{ROI} are manually drawn. Note that, with the \ac{KSF} module we now select the top-$1$ tumor and top-$1$ slice from automatic detection, as opposed to the manual lesion tracing in \Sec~\ref{sec:tumor_charac}. We also compare against variants without \ac{KSF}, meaning that we just select the lesion candidate with the highest detection score. As \Tab~\ref{table:pipeline} shows, both CenterNet-3D and \ac{KSF} contribute significant performance improvements, validating our choices. While performance gaps remain compared to a manual approach, these results demonstrate that our lesion characterization pipeline can provide an effective \textit{fully-automated} solution. Given the challenging nature of our dataset and the coverage/size of our patient population, the results are highly encouraging. 

\subsection{Comparison with Human Physicians}
\label{sec:unified}



Even though it is not a direct comparison, we can put our results in context by comparing against the reported human reader performance. The largest study-to-date is a multicenter trial ($544$ nodules from $381$ patients), that comprehensively evaluated physician performance using dynamic contrast CT, along with MRI and contrast-enhanced ultrasound~\cite{Aube_2017}. Reported F1 scores for differentiating \ac{HCC} vs. others is $0.769$~\cite{Aube_2017}. Our system's performance for \ac{HCC} vs. others compares well, with an F1 score of $0.797$ (manual+SaDT) and $0.763$ (fully automated). However, the patient populations do differ. Aubé \etal{} focus on $1$-$3$ centimeter large tumors~\cite{Aube_2017}, whereas we have no size restriction. On the other hand, our patient selection is arguably more challenging because we only select  pathologically confirmed cases. This means that our patient instances are biased towards studies with high diagnostic uncertainties by radiology alone, while Aubé \etal{}~\cite{Aube_2017} sample from a more general population of cirrhotic patients with nodules. Finally, reported human reader performance can vary, \eg{} another comprehensive human reader study with similar selection criteria report an HCC vs. others F1 score of $0.690$~\cite{khalili2011optimization}. It should also be noted that an important benefit of computer-aided diagnostic systems is not necessarily that they should ``match'' human readers. Instead, it is that they may provide an ``objective'' piece of evidence that can be reliably weighed within clinical decision making. Thus, these comparative numbers above suggest that our system is a promising avenue for helping differentiate liver lesions.

\section{Discussion}
We present and evaluate both semi-automated and fully-automated cancer imaging diagnosis frameworks for liver tumor characterization from multi-phase dynamic CT studies. Extensive experiments are conducted on a semi-supervisedly curated clinical dataset of $1305$ multi-phase CT studies of 5220  scans, all with pathology-proven diagnoses. This patient population represents some of the most challenging lesion characterization cases which cannot confidently be resolved using radiological imaging by human readers assisted with patient meta-information, \eg{}, blood work, age, etc. Our quantitative results validate that the proposed 3D tumor detection, data curation, \ac{KSF}, and \ac{SaDT} modules all provide significant performance contributions. 

To the best of our knowledge, we are the first to represent the task of liver tumor characterization using deep dynamic image texture learning corresponding to multi-phase CT images, rather than treating tumor characterization as a generic object recognition application. While promising performance has been achieved on our challenging patient dataset, there is ample future work. For one, further data collection to address the imbalanced training classes, \eg{}, $\approx$ 77 \% are HCC and there is a paucity of ICC patients, would certainly be of benefit. Multi-center data collection could be a viable strategy, with our curation pipeline being a key component for scalable generalization. Another important direction is to incorporate complementary radiomic features, \eg{}, shape and size~\cite{Aerts2018Artificial,siegel2018assessment}. Finally, it is critical to model prediction confidences~\cite{ghesu2019quantifying} and increase the decision explainability by more explicitly describing the dynamic contrast patterns triggering certain classifications. These efforts can further enable our system to more reliably provide diagnoses in assisting physicians. These directions should push progress even further toward minimizing the use of potentially risky and painful invasive biopsy and surgical procedures.


In this study, a subset of $304$ patient studies have been manually annotated with the primary tumor location rather than the entire cohort. Collecting more annotations was hindered by the labor costs to obtain clinical-quality pixel-level gold standard annotations, which would have required our collaborating physicians to confirm/annotate each tumor in 3D/4D imaging data based on both radiological and pathological reports. Among the fully-labelled studies, $196$ were used as training data, while $108$ were used for testing.  The $108$ studies used for testing is larger than many clinical studies, e.g., the $101$ nodules from $84$ patients used by Khalili et al.~\cite{khalili2011optimization}, and matches or exceeds other computer-aided diagnosis studies~\cite{yasaka2018deep,freeman2006optimizing,kim2007peripheral}. Meanwhile, our training dataset is weakly annotated, which allowed for its comparatively larger size. Nevertheless, our overall novel technical framework and all method components are adequately validated which may inspire new directions on tackling cancer imaging characterization.


\section{Conclusion}
We presented a novel multi-stage deep dynamic texture learning framework for liver tumor characterization under multi-phase dynamic CT. We employ a 3D non-isotropic anchor-free  lesion detection network for both a highly effective semi-automated tumor harvesting procedure in training and for tumor \ac{ROI} extraction during inference. A \ac{KSF} module further selects reliable key slices in inference. With the \acp{ROI} extracted, we approach liver tumor differentiation from a deep texture learning perspective using the proposed \acf{SaDT} network. Extensive quantitative experiments are conducted on a curated dataset of $1305$ multi-phase CT\ patient studies (each with four CT imaging scans and the pathologically confirmed tumor diagnosis result as his/her patient-level label), to sufficiently demonstrate that our 3D lesion detection, data curation, \acf{KSF}, and \acf{SaDT} models all contribute to the final performance. Our \ac{CT} analysis system compares very favorably to reported clinician performance when they use both CT and MRI protocols~\cite{khalili2011optimization}.


\ifCLASSOPTIONcaptionsoff
  \newpage
\fi



\bibliographystyle{IEEEtran}
\bibliography{main1}
%






\end{document}